\RequirePackage{fix-cm}
\documentclass[smallextended]{svjour3}       
\smartqed  
\usepackage{graphicx}
\usepackage[compact]{titlesec}  
\usepackage{natbib}
\usepackage[english]{babel}
\usepackage[utf8]{inputenc}
\usepackage{amsmath, comment}
\usepackage{lscape}
\usepackage{graphicx}
\usepackage[caption=false,font=footnotesize]{subfig}
\usepackage{amssymb}
\usepackage{pifont}
\usepackage{xcolor}
\usepackage{geometry}
\usepackage{multirow}
\usepackage{mathtools}
\usepackage{stfloats}
\usepackage{textcomp}
\usepackage{makecell}
\addto\captionsenglish{}

\begin{document}
%
\title{Bond Graph Modelling and Simulation of Pneumatic Soft Actuator\thanks{Supported by organization DST, India and NRF, South Korea.}} 

%
%
\author{Garima Bhandari \and
Pushparaj Mani Pathak \and
Jung-Min Yang}
\authorrunning{G.Bhandari et al.}
%
\institute{Indian Institute of Technology Roorkee, Haridwar-24766, Uttarakhand, India \and
Kyungpook National University, 80, Daehak-ro, Buk-gu, Daegu, South Korea\\
\email{gsoharu@me.iitr.ac.in, pushparaj.pathak@me.iitr.ac.in}\\
\email{jmyang@ee.knu.ac.kr}}
\maketitle              
\begin{abstract}
This paper presents the design and dynamic modelling of a soft pneumatic actuator that can be used to mimic snake or worm-like locomotion. The bond graph technique is used to derive the dynamics of the actuator. To validate the accuracy of the derived dynamic model, we conduct numerical simulations using 20-sim\textregistered~ software. Experimental results demonstrate that the soft actuator achieves bi-directional bending and linear displacement, which is essential for mimicking snake or worm-like locomotion.

\end{abstract}
%
%
%

\section{Introduction}
In nature many limbless reptiles with slender bodies use their body's flexibility and frictional properties to move around and overcome obstacles in environment. This has inspired new field of soft robots which mimic the motions of natural creatures like a snake, elephant trunk, etc. The rigid robots will require an infinite number of joints to realize the flexibility to mimic the behavior of snake locomotion, which is not an energy-efficient solution. Whereas soft robots have an infinite passive degree of freedom (DoFs), enabling them to be passively deformed when interacting with environments under simple actuation. Another advantage of soft robots over conventional rigid robots is safe interaction with humans and adaptability to complex and uncertain environment at low-cost \citep{Laschieaah3690}.

Several soft robots perform various movements as bending, rotation, extending/contracting, and twisting using multiple techniques and materials, including shape memory alloys, shape morphing polymers, dielectric elastomers, tendon drive, piezoelectric actuation, and fluid power \citep{8258859,JOUR,Natarajan,8725920}. Amongst these, pneumatically-driven are found most advantageous because of their large deformation/force, good power-to-weight ratio, and low manufacturing cost\citep{6584080}. They have gained popularity as a fiber-reinforced soft bending actuator (FRSBA) \citep{6630851}, but the use of viscoelastic material makes soft robots a highly non-linear system. The non-linearity of the dynamics of the soft robots increases the complexity of control \citep{act7030048}, due to which most control strategies are open-loop in the case of soft robots \citep{Huang}.

Few studies have attempted to implement close loop control by dynamically controlling the fluid pressure or mass inside the robot body \citep{10.1109/IROS.2016.7759139, 7914755}. Some have ignored dynamics completely and tried model-free control methods or training methods \citep{7110394, Wu}. But ignoring dynamics does not guarantee accurate control performance. Further, few studies implemented theoretical kinematics and dynamics models for controller design. These studies included various approaches, including finite element method(FEM), constant curvature method, concentrated mass method, and Euler-Lagrange method  \citep{7559710, doi:10.1177/0278364915587926, Thor}. The dynamics provided by before mentioned methods are ambiguous, complex, and difficult to analyze, thus making controller design difficult. Further few studies implemented empirical methods such as Jacobian method \citep{8115276}, visual servo control \citep{7989648}, neural network, sliding mode control and adaptive control \citep{10.3389/frobt.2019.00022, 7827097}. These studies provide a simpler approach toward system modeling and controller design than the strategies involving more complex theoretical models. However, the nonlinear behaviors are not perfectly described, and the validities are limited to the specified experimental conditions \citep{9162458}.

The paper presents a simple soft actuator design based on has bi-directional bending and stretching/contracting capabilities due to which it can have snake or worm-like movements. The actuator can be used in various applications as a manipulator or mobile robot. Further, a dynamic model of the soft actuator has been derived from the bond graph modelling approach. The proposed modeling formalism can accommodate major dynamic parameters of the soft actuator and system non-linearities. The robot has been controlled using PD controller. The performance of the soft actuator modelling is validated through numerical simulations where the soft robot emulates bi-directional bending and extension/contraction as desired by the controller input.

The rest of this paper is structured as follows. Section 2 presents the design and modelling of a soft actuator in framework of bond graph technique. Section 3 presents detailed numerical simulation results to validate proposed dynamics with close-loop PD control implemented. Finally, Section 4 concludes the paper.

\section{Dynamic Modelling}

We will discuss the dynamics of the soft actuator in this section. To understand the dynamics it is important to understand the working and design of the soft actuator. The following sections explains in detail the design and bond graph modelling of the actuator. 
\subsection{Design of Soft Actuator}
\begin{figure}[ht]
	\centering
	\includegraphics[width=0.45\textwidth]{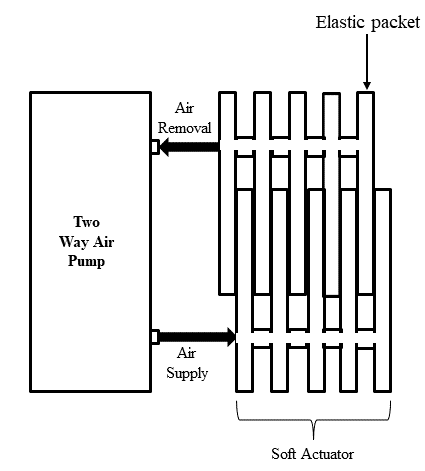}
\caption{Schematic representation of working concept.}
\label{working concept}
\end{figure}

The design of the soft actuator is bellow-like. The soft actuator has packets made of elastic material joined together, as shown in Fig.~\ref{working concept}.

The elastic packets inflate and deflate like balloons when air is supplied or removed from them. The two-way air pump can supply and remove air;  the connection of air pump to soft actuator is as shown in Fig.~\ref{working concept}. The bidirectional bending is emulated by alternating supplying and removing air from each side of the soft actuator. The actuator bends in the direction of the side from where the pump removes air.

\begin{figure}[ht]
	\centering
\subfloat[]	{\includegraphics[width=0.5\textwidth]{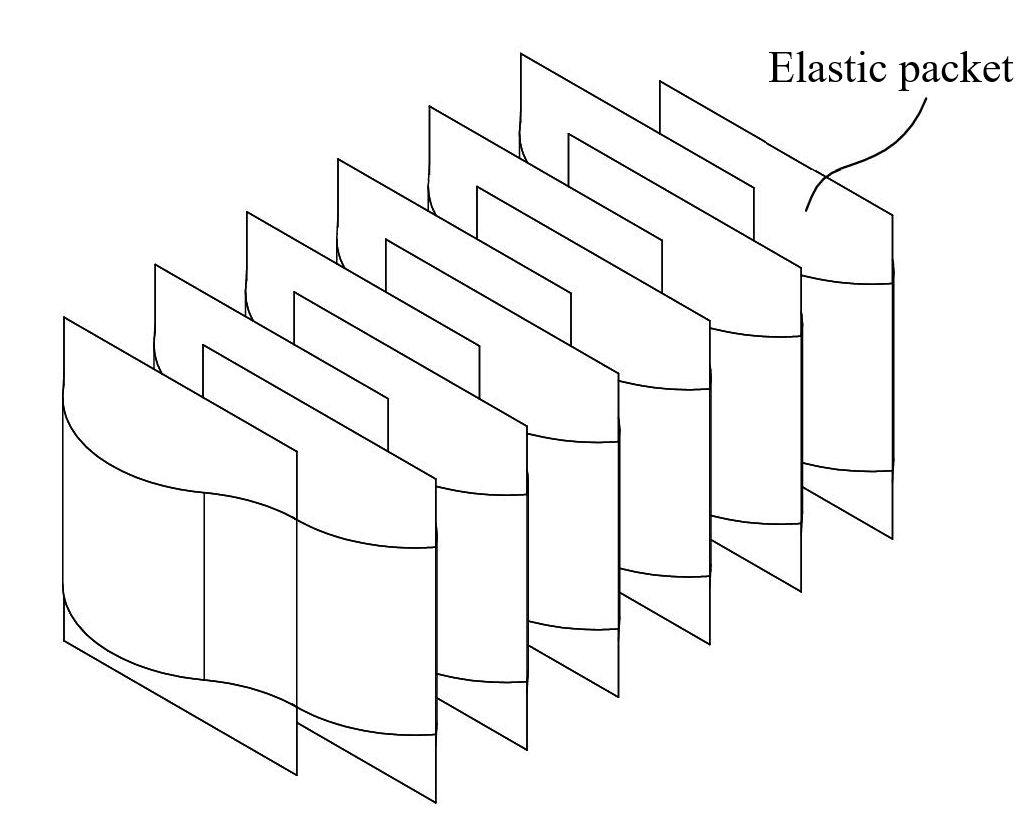}
		\label{solid}}
\subfloat[]{\includegraphics[width=0.5\textwidth]{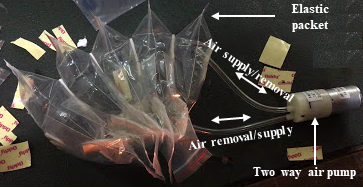}
		\label{soft actuator}}	
\caption{Soft actuator's (a) solid model, and (b) pictorial representation of its working concept.} \label{concept}
\end{figure}

 The micro-controller controls the alternating air supply and removal of air for bidirectional bending. We can attach various sensors to the soft actuator to measure the pressure, extension, and bending angle and employ them to control the soft actuator. Fig.~\ref{solid}\textendash\ref{soft actuator} shows solid model and the pictorial representation of the working concept of the soft actuator. Now with a clear understanding of working and control of soft actuator, we will model the bond graph to derive the dynamics in the next section.

\subsection{Bond Graph Modelling of Soft Actuator}
\begin{figure}[httb]
  \centering
  \includegraphics[width=0.5\textwidth]{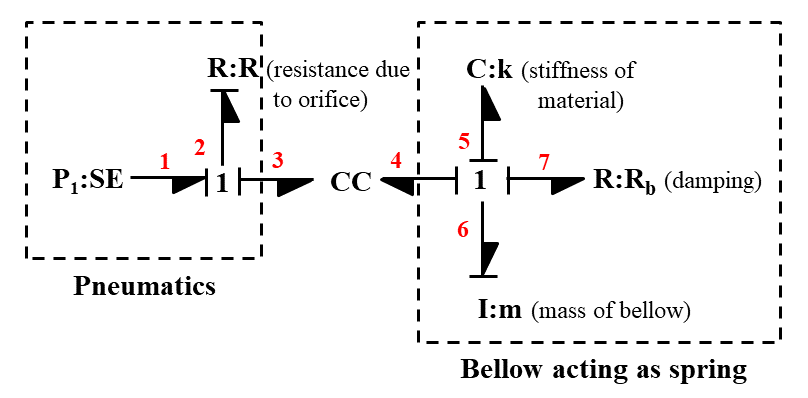}
  \caption {Bond graph of the bellow.}
  \label{bellow}
  \end{figure}

The bond graph modelling technique reduces mathematical complexities related to highly non-linear system, such as soft actuator presented in this paper. Since our actuator design is bellow-like, so we model each elastic packet as a bellow. For the basic understanding of bond graph model, we explain Fig.~\ref{bellow}. in detail.

The source of effort element \textbf{(SE)},is the pressure ( \textit{$P_1$}) supplied to the elastic packet from the air pump. The resistance element \textbf{(R)}, is the resistance the pneumatic flow faces when it enters through the orifice in the elastic packet. We calculate value of resistance as:  
\begin{equation}
R = \frac{{\sqrt {P_1-P_2} }}{{{C_d}D}}
\end{equation}
where, \textit{$C_d$}, is the orifice coefficient of discharge, and \textit{D} is diameter of orifice.

We have assumed extension and contraction of soft actuator synonymous with that of spring for the ease of modelling; with this assumption, we infer that the effort from the pneumatic domain is causing displacement in the mechanical part. We use C-field element \textbf{(CC)} in cases like these where the efforts and displacements are in a different domain but interrelated. Thus we have two capacitance values in this system, \textit{$C_1$},  due to change in volume of the elastic packet and \textit{$C_2$}, due to compressibility  of air \citep{Bolton}. We can calculate their value as follows:
\begin{equation}
\begin{array}{l}
{C_1} = \frac{{{A^2}}}{k}\\ \\
{C_2} = \frac{{Ax}}{\rho {RT}}
\end{array}
\end{equation}
where \textit{A} and \textit{k} are the area and stiffness of elastic packet respectively, \textit{x} is the extension in the elastic packet after air supply, $\rho$ is the density of air, R is gas constant, and \textit{T} is the temperature of air supplied.

The resistance element \textbf{(R)} represented as \textit{$R_b$} is damping in elastic packet  and inductance element (\textbf{I}), is its mass (\textit{m}). The dynamic equations we get after bond graph modelling are:
\begin{equation}
R({C_1} + {C_2})\frac{{d{P_2}}}{{dt}} + {P_2} = {P_1}
\end{equation}
\begin{equation}
m\ddot x + {R_b}\dot x + kx = {P_2}A
\end{equation}

where \textit{$P_2$} is the pneumatic pressure inside the elastic packet, which is given by effort in bond number 3 in Fig.~\ref{bellow}.

\begin{figure}[httb]
  \centering
  \includegraphics[width=0.6\textwidth]{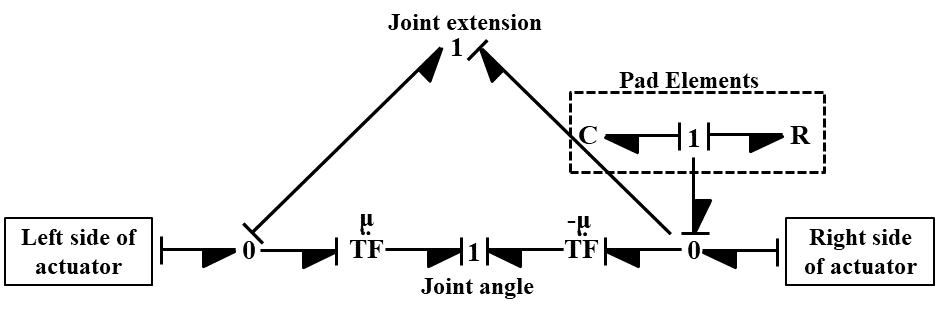}
  \caption{Bond graph of the actuator.}
  \label{actuator}
  \end{figure}

The bond graph model of elastic packet is then used to make the both sides of the actuator as shown in Fig.~\ref{actuator}. 
We have implemented half car model for bidirectional bending of the actuator. Since we want to alternately supply and remove air from each side of actuator , the pressure \textit{$P_1$} is provides as:
\begin{description}
\item[Left side of actuator]
\end{description}
\begin{equation}
\label{leftside}
\begin{array}{l}
{P_1} = {k_p}({x_{Lref}} - {x_L}) + {k_d}({{\dot x}_{Lref}} - {{\dot x}_L})\\ \\where, {x_{Lref}} = x\sin (wt)
\end{array}
\end{equation}
\begin{description}
\item[Right side of actuator]
\end{description}
\begin{equation}
\label{rightside}
\begin{array}{l}
{P_1} = {k_p}({x_{Rref}} - {x_R}) + {k_d}({{\dot x}_{Rref}} - {{\dot x}_R})\\ \\
where,{x_{Rref}} = x\sin (wt+\pi)
\end{array}
\end{equation}
$k_p$ and $k_d$ are proportional and derivative control gains, whereas $x_L$ and $x_R$ are actual extension or contraction of left and right side of the actuator.

From Eqs.\ref{leftside}-\ref{rightside} we infer that we alternately supply air to the left side of the actuator and remove air from its right side, thus making it bend in both directions. We can also achieve simple extension and contraction by providing air supply as in Eq.\ref{leftside} on both sides of the actuator.

\section{Numerical simulations}
After creating bond graph model of actuator, we conduct numerical simulations to validate its accuracy in the framework of 20-sim\textregistered~ (ver.~4.7) \citep{20sim}. Table \ref{System parameters} summarizes the system parameters of actuator used in the numerical experiment.
\begin{table}[ht]
\renewcommand{\arraystretch}{1.5}
\footnotesize
\caption{System parameters used in numerical experiment.}\label{System parameters}
\begin{center}
\begin{tabular}{cc|cc}
\Xhline{3\arrayrulewidth}
Parameter & Value & Parameter & Value\\
\hline
\textit{m}  &  0.015 kg & \textit{x} & 0.3 m \\
$R_b$ &  0.4 kg/sec & $\rho$ & 1.225 kg/$m^3$ \\
\textit{k} & 350 N/m & R & 8.31451 J/K mol\\
$C_d$ & 0.8 & \textit{T} & 300 K\\
\textit{D} & 0.008 m& $k_p$ & 40 \\
\textit{A} & 0.0096 $m^2$ & $k_d$ & 10 \\
$\mu$ & 2.5 & $\omega$ & 1 rad/sec\\
\Xhline{3\arrayrulewidth}
\end{tabular}
\end{center}
\end{table}

\begin{figure}[ht]
	\centering
\subfloat[]	{\includegraphics[width=0.52\textwidth]{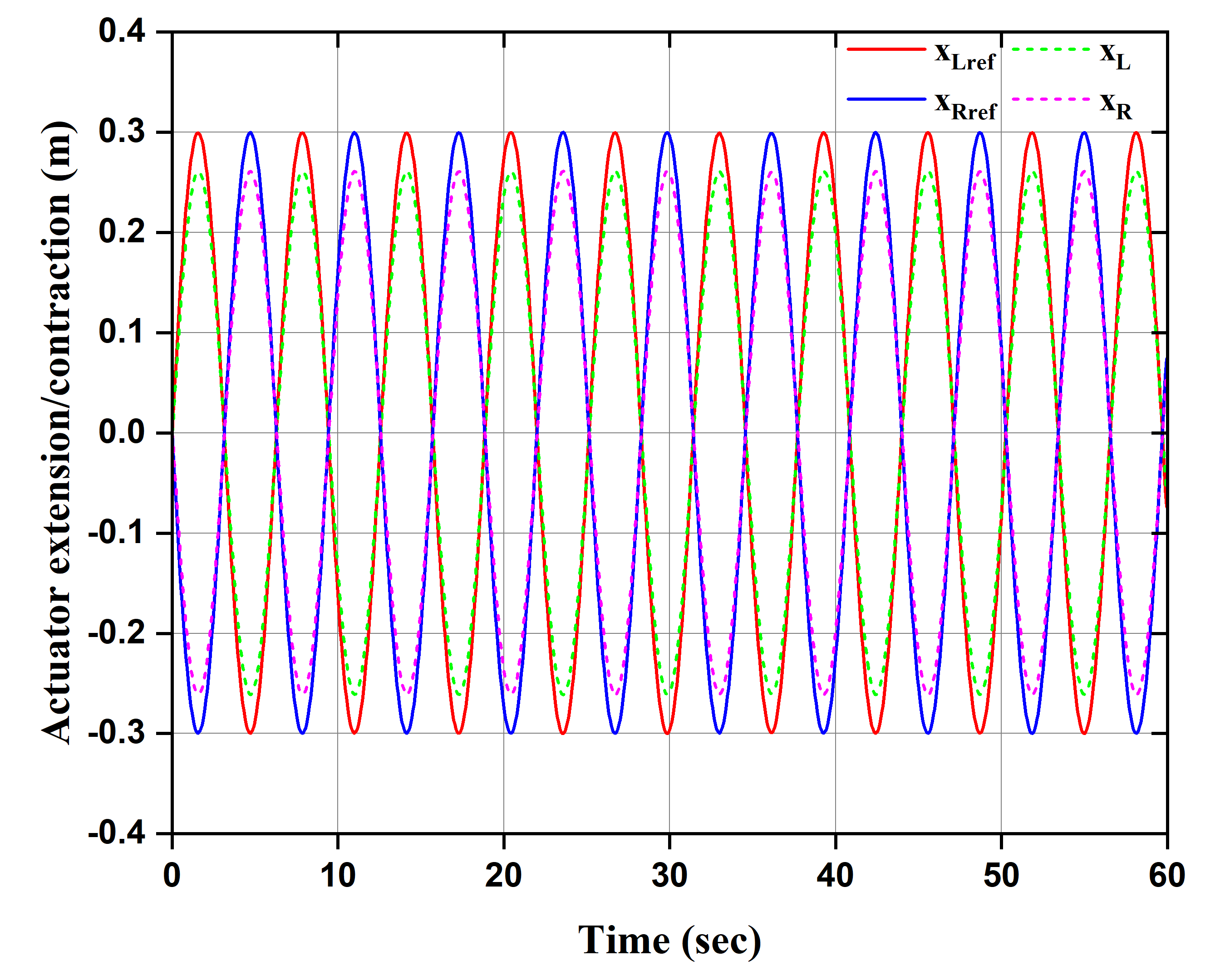}
		\label{extension}}
\subfloat[]{\includegraphics[width=0.57\textwidth]{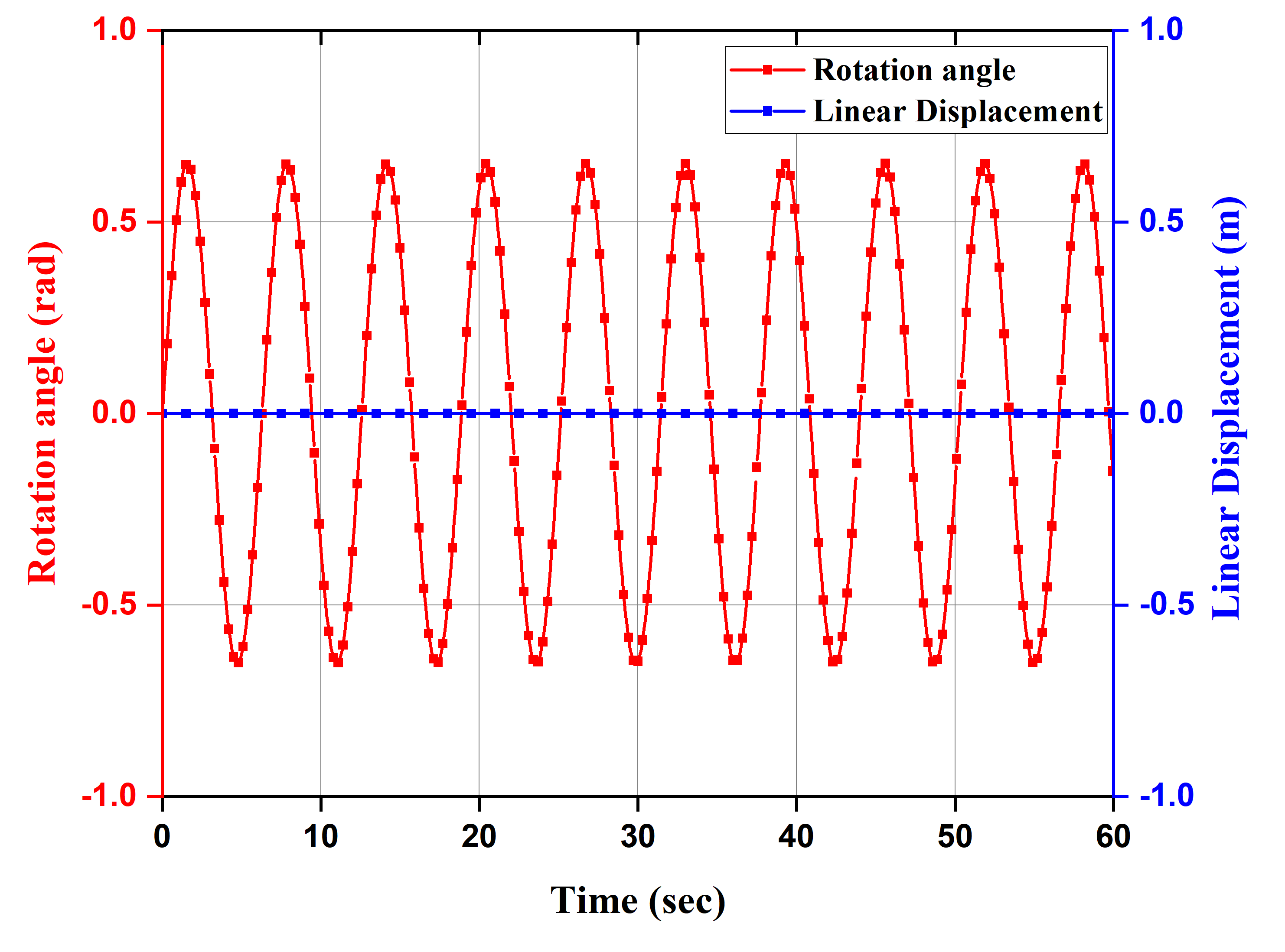}
		\label{jointangle}}
		\hfil
\subfloat[]	{\includegraphics[width=0.52\textwidth]{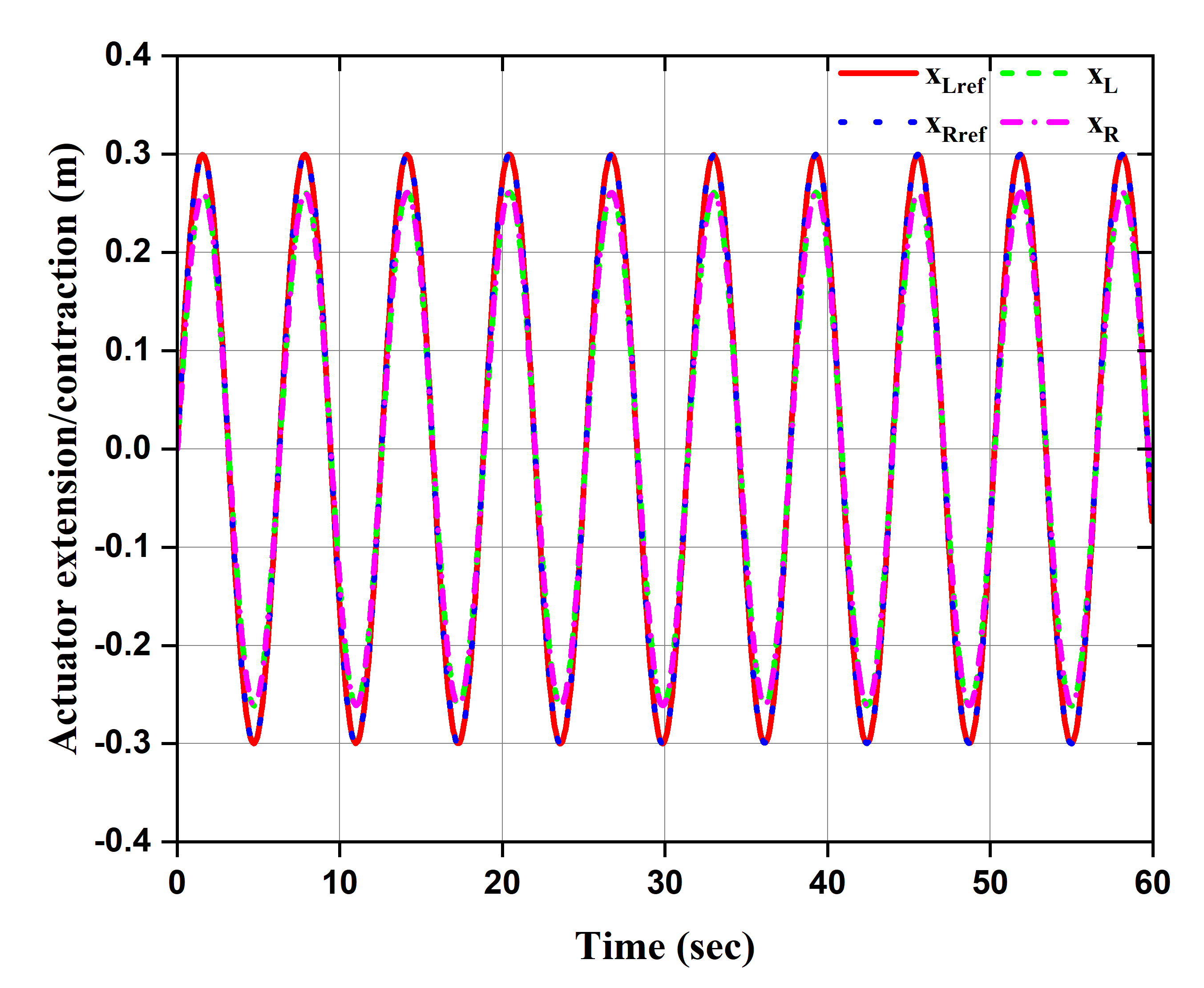}
		\label{extension1}}
\subfloat[]{\includegraphics[width=0.57\textwidth]{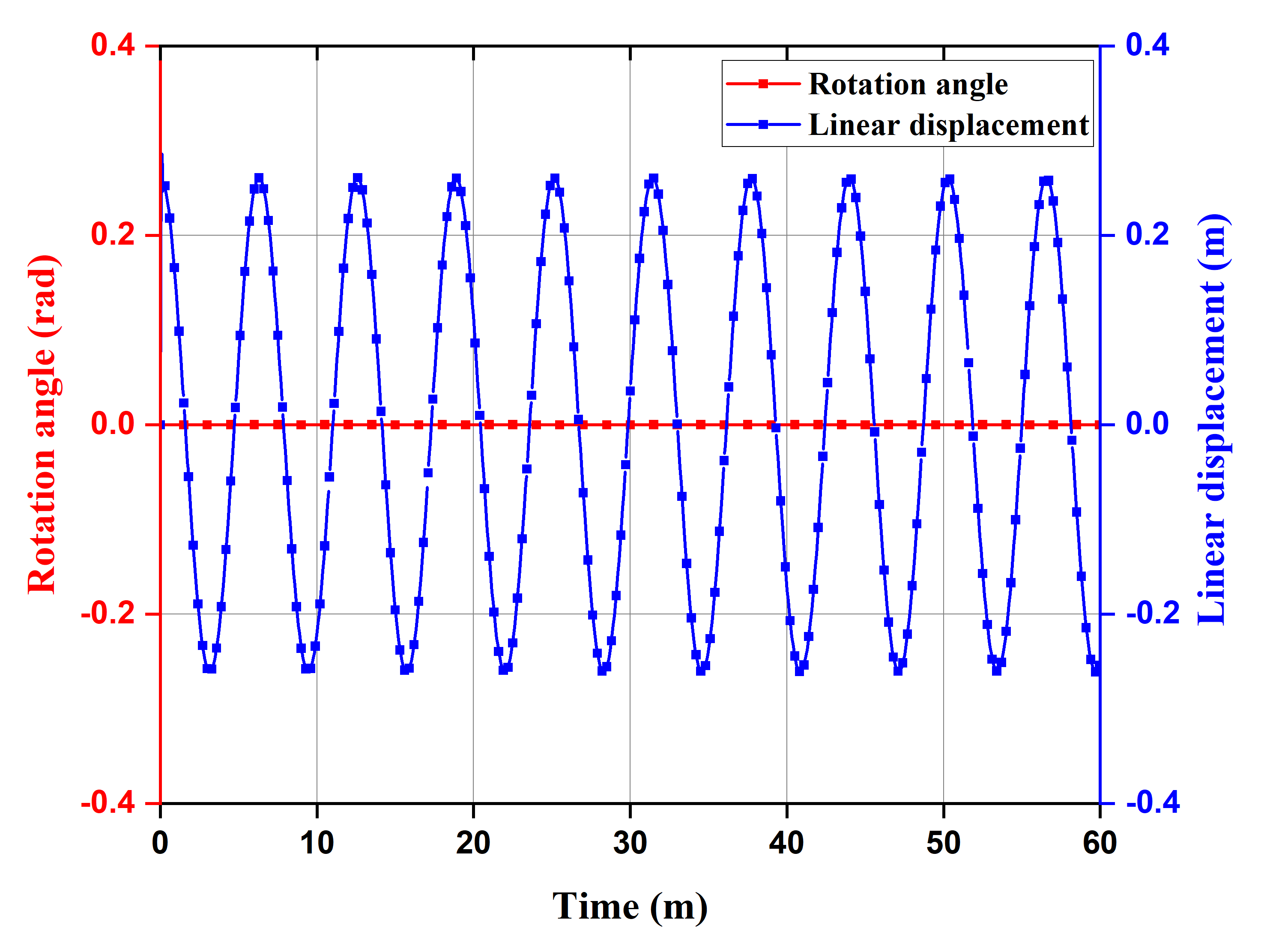}
		\label{jointextension}}		
\caption{The behaviour of actuator when (a) air is removed and supplied alternately, (b) rotation and extension due to alternate air supply, (c) air is removed and supplied simultaneously, and (d)  rotation and extension due to simultaneous air supply. } \label{actuator extension}
\end{figure}

Figs.~\ref{extension} and \ref{jointangle} shows results for bi-directional bending of actuator. We can see in, Fig.~\ref{extension}, each side of actuator extends and contracts alternately according to $x_{ref}$ for each side. Fig.~\ref{jointangle} shows the result for bi-directional bending and rotation angle achieved. We can see there is no linear displacement in this case. 

The extension and contraction of both sides of the actuator for linear displacement of the actuator are drawn in Fig.~\ref{extension1}. In this case, we can see both sides of the actuator are expanding and contracting simultaneously. Fig.~\ref{jointextension} demonstrates that the actuator has only linear displacement when each side expands and contracts simultaneously.

\begin{figure}[ht]
	\centering
\subfloat[]	{\includegraphics[width=0.5\textwidth]{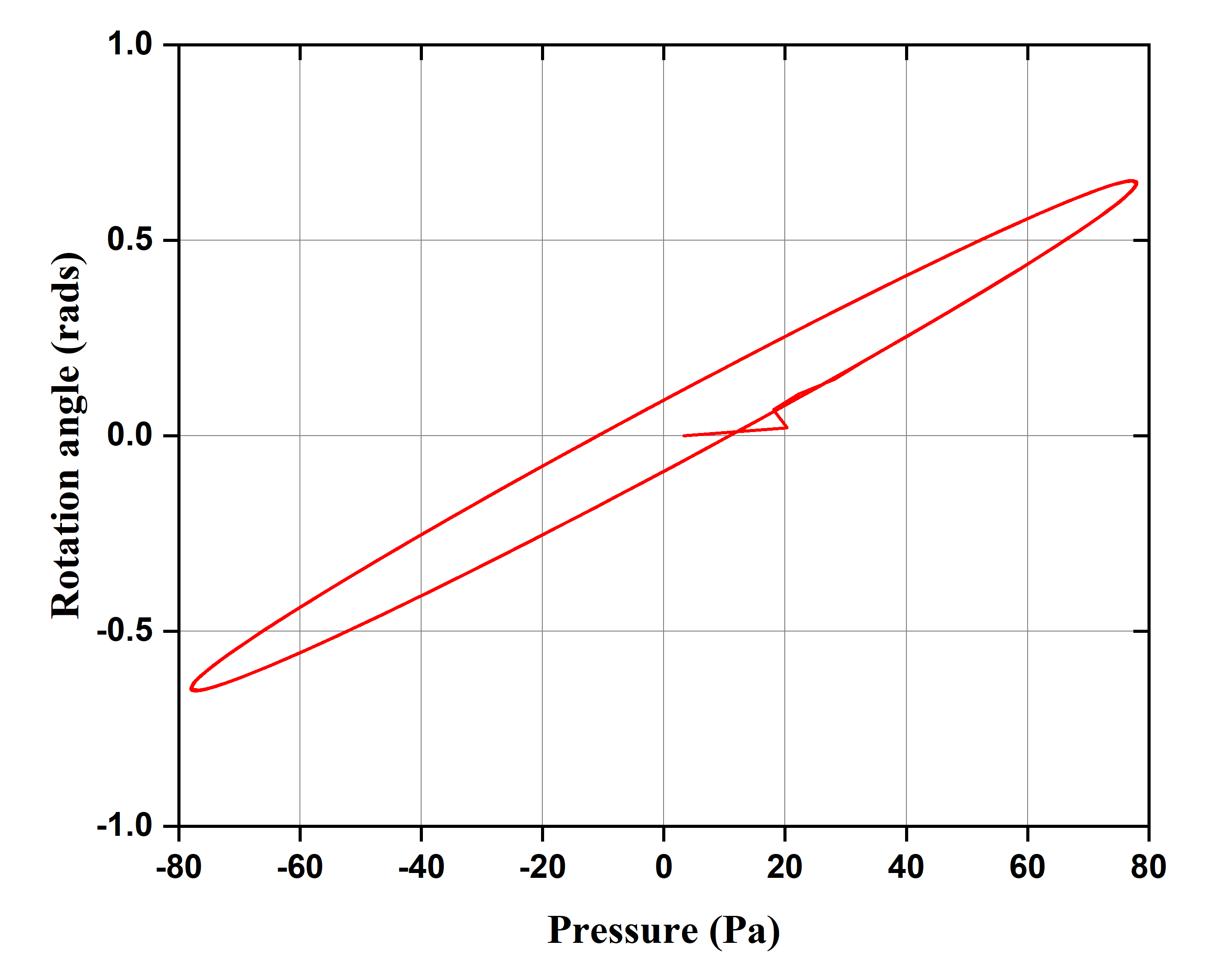}
		\label{rp}}
\subfloat[]{\includegraphics[width=0.5\textwidth]{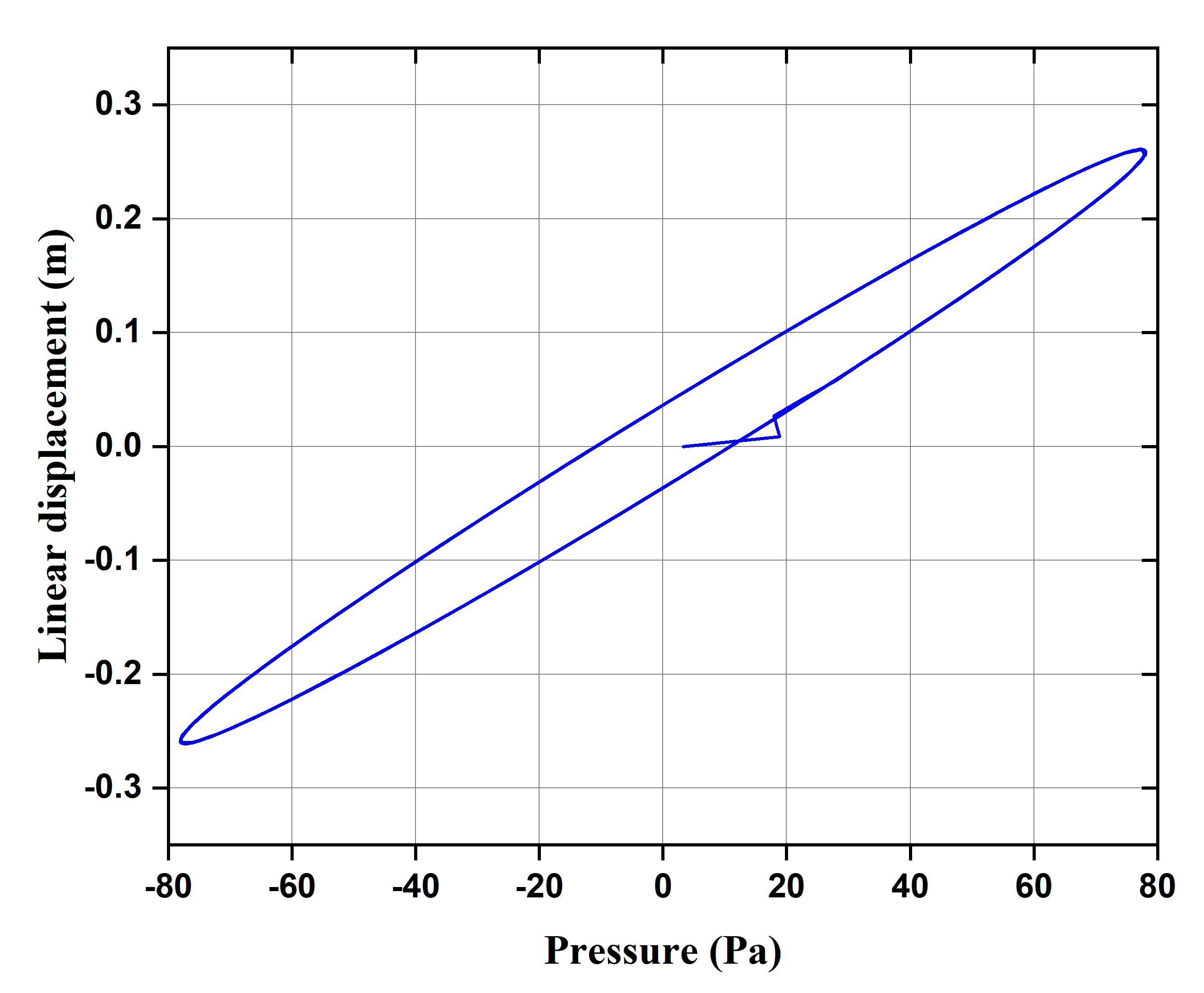}
		\label{ep}}
		\hfil
\subfloat[]	{\includegraphics[width=0.5\textwidth]{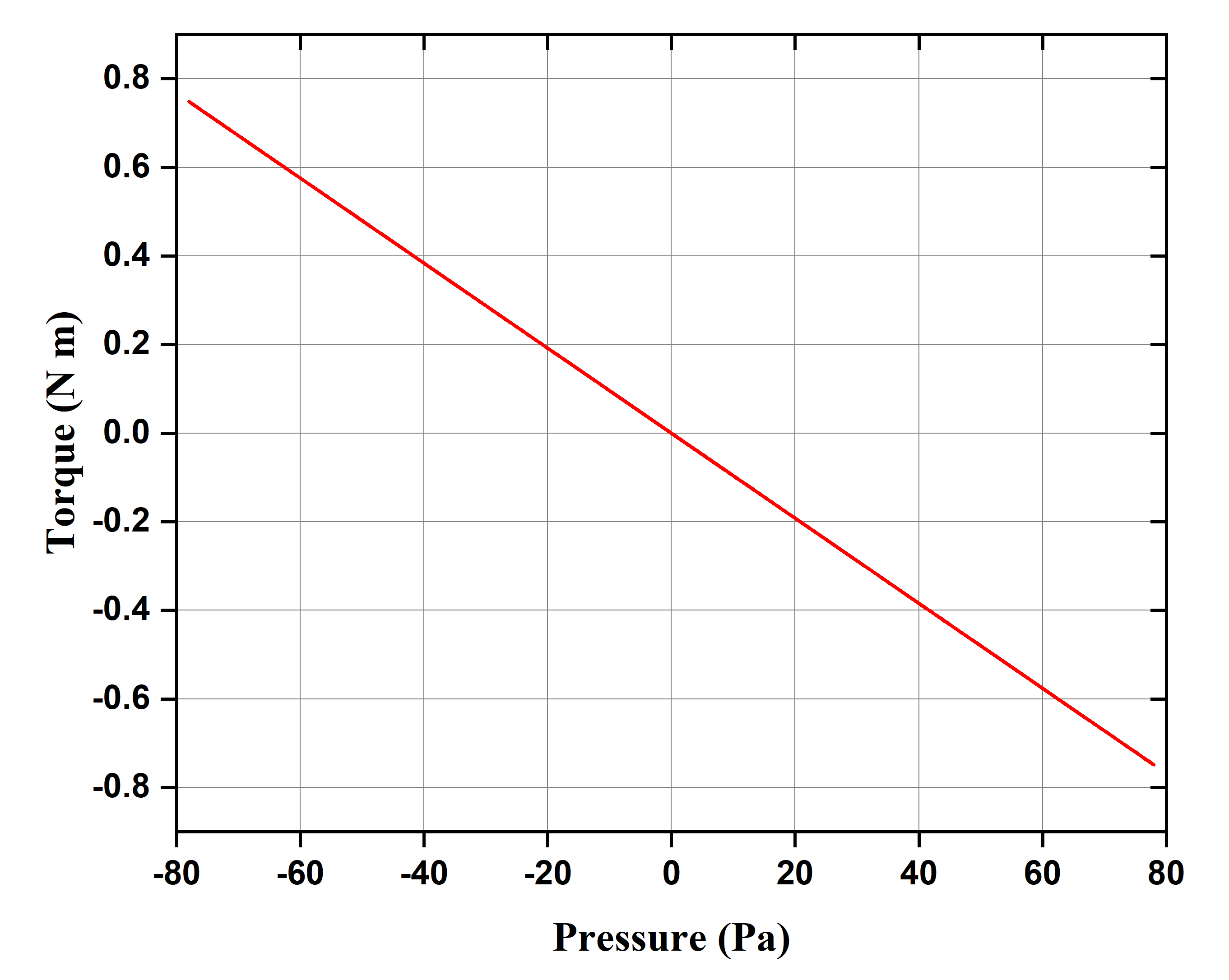}
		\label{tp}}
\caption{Performance of actuator in terms of (a) rotation, (b) linear displacement, and (c) torque obtained for the controlled output.} \label{pressure}
\end{figure}

As explained in section.~2.2 controller output is pressure, $P_1$, supplied to actuator ( see Eqs.~\ref{leftside}$\textendash$\ref{rightside}). Fig.~\ref{pressure} shows the results for rotation angle, linear displacement and torque generated by actuator for the controlled pressure supply.

\clearpage

\section{Conclusion}

The paper presented a design and dynamic modelling of the soft actuator, which can be used in the varied application. The dynamics developed in this paper will be the basis for the complete dynamic modelling of the manipulator or robot created using our actuator. The conventional PD control is used in this paper, but we can use superior control methods since we have an accurate dynamic model. It would be interesting for future studies to achieve other important objectives such as path planning and obstacle avoidance for a soft robot made from the soft actuator.
 
 \section*{Acknowledgment}
This work was supported in part by Indo-Korea JNC Program of Department of Science and Technology (DST), Government of India (INT/Korea/JNC/Robotics, dated 23-03-2018), in part by the National Research Foundation of Korea (NRF-2017K1A3A1A68072072).

\bibliography{ref}

\section*{Nomenclature} \vspace{0.1cm}
\noindent
$\mu$\space\space\space\space Transformer coefficient in bond graph model\\ \\
$\rho$\space\space\space\space Density of air supplied by air pump\\ \\
$\omega$\space\space\space\space Angular frequency for desired extension/contraction\\ \\
A\space\space\space\space Area of elastic packet\\ \\
$C_d$\space\space\space Orifice coefficient of discharge\\ \\
D\space\space\space\space Diameter of orifice in elastic packet for air supply/removal\\ \\
k\space\space\space\space Stiffness of elastic packet\\ \\
$k_d$\space\space\space Derivative gain constant\\ \\
$k_p$\space\space\space Proportional gain constant\\ \\
m\space\space\space\space Mass of elastic packet\\ \\
R\space\space\space\space Gas constant\\ \\
$R_b$\space\space\space Damping in elastic packet\\ \\
T\space\space\space\space Temperature of air supplied\\ \\
x\space\space\space\space Amplitude of desired reference trajectory

\end{document}